# HUMANOID ROBOT WITH VISION RECOGNITION CONTROL SYSTEM

**Cosmin Başca, Mihai Taloş, Remus Brad**

Abstract: This paper presents a solution to controlling humanoid robotic systems. The robot can be programmed to execute certain complex actions based on basic motion primitives. The humanoid robot is programmed using a PC. The software running on the PC can obtain at any given moment information about the state of the robot, or it can program the robot to execute a different action, providing the possibility of implementing a complex behavior. We want to provide the robotic system the ability to understand more on the external real world. In this paper we describe a method for detecting ellipses in real world images using the Randomized Hough Transform with Result Clustering. Real world images are preprocessed – noise reduction, greyscale transform, edge detection and finaly binarization – in order to be processed by the actual ellipse detector. After all the ellipses are detected a post processing phase clusters the results.

## 1. INTRODUCTION

Humanoid robots are the best solution for Human Computer Interaction (HCI) based research. Unlike other type of robotic systems (e.g. wheeled robots) humanoid robots can use other forms of intuitive communication such as body language, mimics and gaze. Humanoid robots are also easily accepted by humans and can easily integrate into environments designed for humans. Up to know a series of humanoid robots have been designed and constructed by prestigious companies like: Sony, Fujitsu, Toyota or Honda. This robotic systems have remarkable technical characteristics but are also very expensive (they can be rented or bought for tens of thousands of dollars), and some of them are only prototypes.

In academic environments it is usually necessary to acquire more than one unit, which is extremely expensive. The solution that we adopted is to augment an already present on the market robotic system. We have chosen WooWee's RoboSapien robot, due to its availability and cheapness compared to other similar systems.

The proposed robotic system is composed of a PC and the robot itself (see Figure 1.).

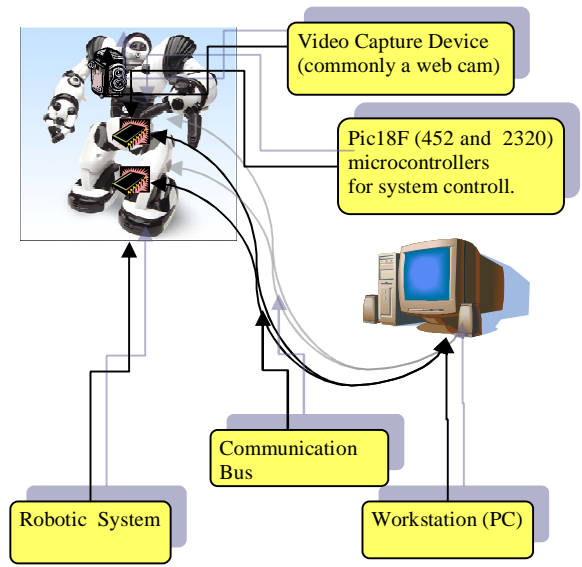

Fig. 1. System overview.

The communication between the two parts of the system is realized through the standard serial port

(RS232). Using an advanced communication protocol, complex actions are forwarded to the robot (e.g. robot initialization, or robot state interrogation). The robot replies with a communication synchronization package, which can contain relevant information about the ongoing action of the robotic system. Feedback is provided via a CMOS – video sensor (a common web cam) and six bumper sensors. The most relevant information about the surrounding environment can be acquired through the use of the CMOS sensor, which provides information under the form of a continuous video stream at up to 25 frames per second, thus being adequate for obstacle detection, real world pattern recognition, motion planning, basically for a greater understanding of the surrounding environment.

In this paper we propose an algorithm for ellipse detection. Ellipses are the most common features that appear in images, (most often circular shapes appear as ellipses due to the projection transform, only spherical objects remain as circles – e.g. balls – when projecting 3D space to 2D space). The Hough technique is particularly useful for computing a global description of a feature(s) (where the number of solution classes need not be known *a priori*), given (possibly noisy) local measurements. The Hough Transform is widely used for parametric curve detection. A *generalized* Hough (Ballard, 1981) transform can be employed in applications where a simple analytic description of a feature(s) is not possible. Due to the computational complexity of the generalized Hough algorithm, this variant of the Hough Transform is rarely used in practice. We will only focus on detecting ellipses using the Hough Transform. The traditional approach for ellipse detection using the Hough technique is similar to line or circle detection (Duda and Hart, 1972). The parametric equation of a line is written as follows:

$$x\cos q + y\sin q = r \quad (1)$$

The Hough Transform is actually a map from *xy* space to the parameter space *rq*. In order to find all the lines in a given image we wold have to solve the given parametric equation for known x and y coordinates. The process of ellipse detection is similar but evidently more complex and more time consuming as the parametric equation would be of the following form:

$$ax^2 + 2bxy + cy^2 + 2dx + 2ey + f = 0 \quad (2)$$

We know have to find all five parameters of the equation in order to detect a valid ellipse. Evidently this approach would require solving the equation for five different points on the ellipse, thus mapping *xy* points to a five dimensional space, (hence having to manage a five dimensional accumulator). This approach is not only memory expensive but also computationally intensive, as the algorithm in its brute search form would have a $O(n^5)$ complexity.

Although there are refined variations of this approach (see Ballard, 1981) these are still expensive considering memory or computational time.

In order to overcome this inconveniencies we have further developed on the idea of Chellali et al., 2003 who use only one dimensional accumulator for ellipse voting, and reducing algorithm complexity.

## 2. BASIC SYSTEM ARHITECTURE

The robot is 34 cm high and 31 cm width. The robots mobile parts are moved using 7 DC motors having 9 degrees of freedom (DOF). One motor is used for left – right tilt, one motor for each foot (moving the foot back and front), one motor for each shoulder (moving the arm up and down) and one motor for each arm (opening the hand to grip and closing it) as can be seen in Figure 2.

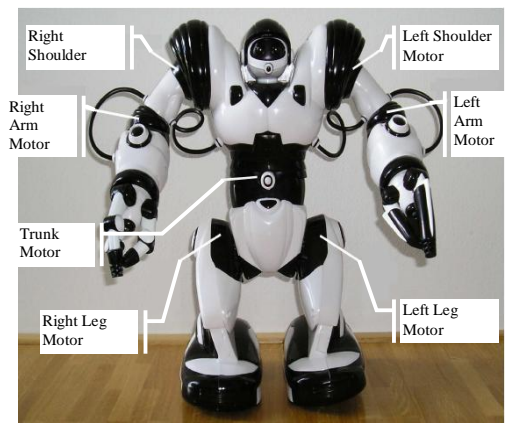

Fig. 2. Robot physical layout

The robot has bumper sensors on both of its legs (in front and behind) and on both of his gripper hands (located on the frontal finger of each hand). Unlike other complex humanoid robots, the robot only uses three of the motors for motion (the trunk motor and both of the leg motors). The idea behind its motion is dynamic walking. The robots upper body swings like a pendulum left and right shifting the weight of the body to one side, allowing the robot to move its opposite foot off the ground. The robot is comprised of two microcontrollers, the interface electronics, and the communication bus with the PC. The DC Motors are powered by a different power source than the one powering the microcontroller (this is necessary in order to eliminate DC voltage fluctuations which could cause the microcontroller to malfunction or to reset). The motors can be driven in both rotation senses.

The microcontroller used for robot motion control is pic18F452 and for camera positioning is used a pic18F2320. The microcontrollers are produced by

Microchip. Using communication interrupts, special interrupt handlers (receive – transmit) are used to implement full – duplex communication. The reception routine checks if the input data is valid – it checks for a synchronization header in the data package. If the synchronization header is not present than the routine resets the state of the reception buffer, the same event occurs when synchronization packages throughout the entire action package are not found or a timeout exception occurs. If the received data comprises a valid action package, than the main program routine is notified and the action package is stored in internal receive buffers for further processing. The routine that realizes the motion of the robot is called based on a timer interrupt, provided by the microcontroller. The main program routine (started at reset) analyzes the received requests from the PC, starts the motion action of the robot, initializes response transmission to PC, and tests if any of the bumper sensors has been activated. A secondary routine is used for middle arm positioning. The arm has a switch which is turned on when the arm reaches its middle position; the switch turns off after the hand leaves the middle position. This secondary routine is input state interrupt based (the interrupt is activated when the microcontrollers input port state changes).

As mentioned before complex motion actions are, accomplished through the use of basic motion primitives. The robot has 21 basic motion primitives as it can be seen in table 1.

TABLE 1: BASIC MOTION PRIMITIVES.

| no. | Motion Primitive | no. | Motion Primitive | no. | Motion Primitive |
|---|---|---|---|---|---|
| 1 | Right Arm Up | 2 | Right Arm Front | 3 | Right Arm Down |
| 4 | Left Arm Up | 5 | Left Arm Front | 6 | Left Arm Down |
| 7 | Right Hand In | 8 | Right Hand Mid | 9 | Right Hand Open |
| 10 | Left Hand In | 11 | Left Hand Mid | 12 | Left Hand Open |
| 13 | Tilt Left | 14 | Tilt Mid | 15 | Tilt Right |
| 16 | Right Leg Front | 17 | Right Leg Mid | 18 | Right Leg Back |
| **19** | Left Leg Front | 20 | Left Leg Mid | 21 | Left Leg Back |

The basic motion primitives can be executed as fast as 1ms at 40 MHz (maximum clock cycle controller speed). The controller has to be powered at 5 V DC in order to operate at 40 MHz, if the power voltage drops below 5 V (if batteries are in use) than the speed of the controller drops to 20 MHz or less.

The robot operates successfully within 40 to 4 MHz microcontroller speed; less than 4 MHz the execution latency exceeds 50ms, resulting in faulty robot control (e.g. the robot can not walk properly – legs do not move symmetrically).

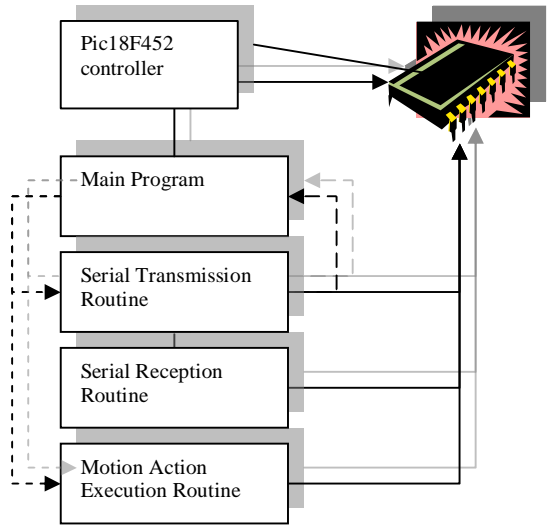

Fig. 3. Program organization

Camera positioning is accomplished both horizontal and vertical providing the robot with a large field of view. The routine used to position the camera highly resembles the main program used for robot motion control, the routine used to execute an action is divided into three parts : one routine for each axis positioning and one for monitoring the global position of the camera.

## 3. PROGRAMABLE API DESCRIPTION

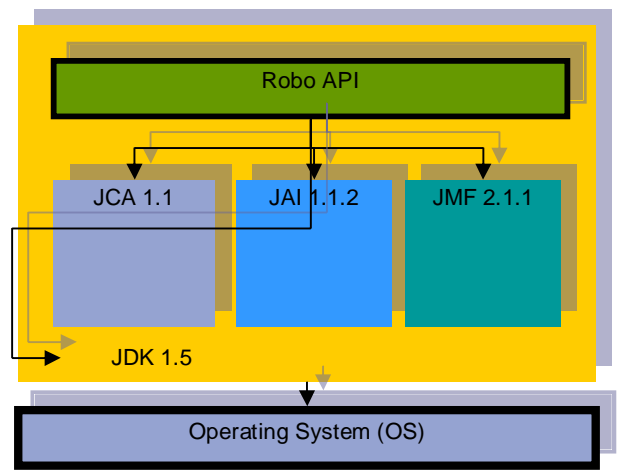

Fig. 4. Programming API architecture description.

Both basic motion primitives and high level actions are stored in a globally available database system called RoboRegistry as it can be seen in Figure 5.

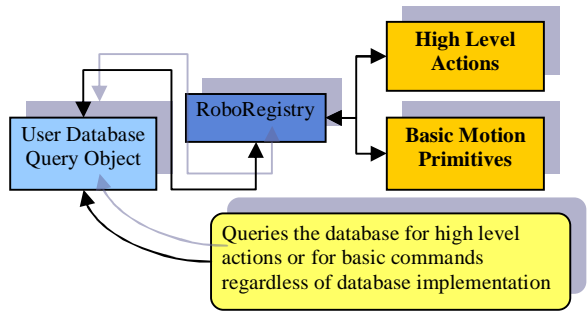

Fig. 5. Database System architecture.

The robotic control is achieved through dedicated control interfaces called *controllers*, the user has to have access to a controller in order to issue requests to the robotic system or to capture feedback from it, as it can be observed in Figure 6.

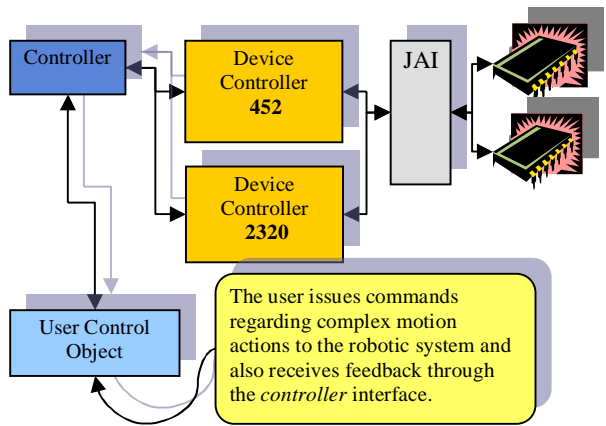

Fig.6. Robotic system high-level control model (The control interface architecture).

To robot is equipped with a CMOS – video sensor (in our case a common web-cam). Through the use of this sensor the system actually receives the most of it's feedback from the external environment. Although the robot has bumper sensors mounted on both it's feet and gripper hands, this sensors provide limited feedback regarding changes in the external environment. The web cam can be used to track motion, detect obstacles, to construct navigation maps, for motion planning, pattern detection (detection of real world objects), human detection etc.

*3.1 Ellipse detection algorithm*

The basic information flow is as described in Figure 7. As it can be seen in Figure 7 the input for the detector block is a binarized image obtained by applying an adaptive threshold (we use the Maximum Variance Threshold as it varies with luminosity) to the gradient image. This step can be further refined, because the detector is highly dependent on its output. All edge pixels are white (the foreground) and the rest are black (background). All contours (features) hence ellipse contours as well should be well defined and noise should be greatly reduced.

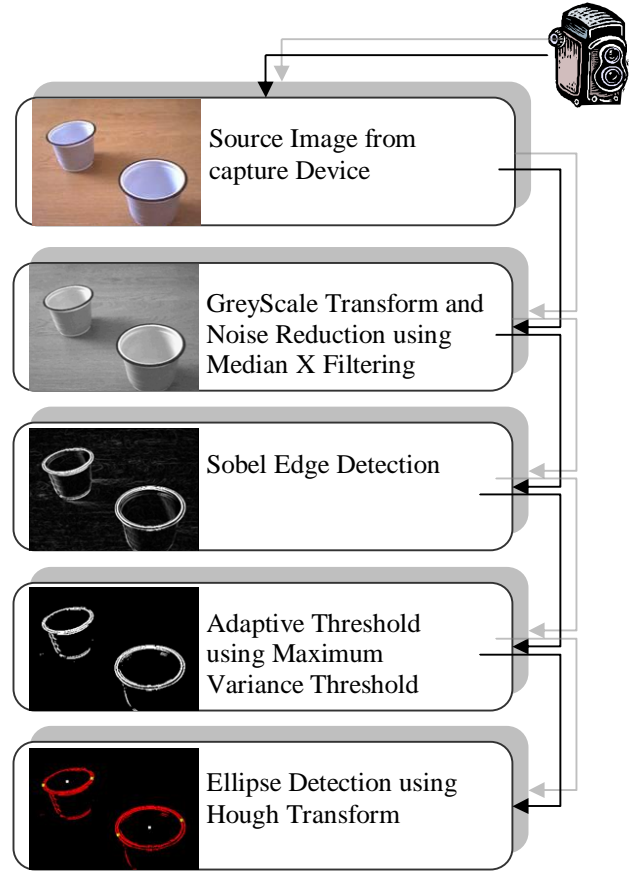

Fig. 7. Processing steps involved in the algorithm.

In order to detect all five parameters of an ellipse only three points are needed, two of which are considered to be the ellipses vertices. We will not detail Chellali et al., 2003 algorithm as it is not the porpoise of this paper, but we will mention the basics of it as it is necessary in order to get an understanding of the modifications we made. Given the two vertices of the ellipse the determination of four of the five ellipse parameters is straightforward using the following formulas:

$$x_0 = \frac{x_1 + x_2}{2} \quad (3)$$

$$y_0 = \frac{y_1 + y_2}{2} \quad (4)$$

$$a = \frac{\sqrt{(x_2 - x_1)^2 + (y_2 - y_1)^2}}{2} \quad (5)$$

$$a = \tan(\frac{y_2 - y_1}{x_2 - x_1}) \quad (6)$$

$x_0$ and $y_0$ being the ellipses center coordinates *a* (please note that a from formula (5) is not the same as a from formula (2)) being half of its major axis and *a* being the orientation of the ellipse.

A third point is needed, in order to determine the fifth parameter of the ellipse – the half length of its minor axis - using the following approximation formulas:

$$b = \sqrt{\frac{a^2 d^2 \sin^2 t}{a^2 - d^2 \cos^2 t}} \quad (7)$$

$$\cos t = \frac{a^2 + d^2 - f^2}{2ad} \quad (8)$$

where d, τ and f can be observed in Figure 8.

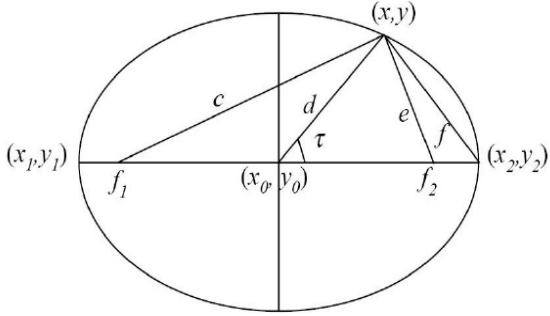

Fig 8. Ellipse geometry, $f_1$ and $f_2$ are the ellipses foci.

Based on formulas (3) to (8) all ellipses in the image can be detected. The algorithm as proposed by Chellali et al., 2003 has a $O(n^3)$ complexity. The complexity of the algorithm can be further reduced by transforming the method into a Randomized Hough Transform (RHT).

In the first step we randomly select *m* pairs of points satisfying the search domain condition (the number of points initially selected is considerably lower than in the original implementation – we only select $m = n*C$ where C = 1, 2, 3,… instead of selecting n*n) thus reducing the complexity to $O(\frac{n^2}{C})$, where as mentioned before C is a constant much smaller than n. The number of selected pairs must be sufficiently large in order to detect all present ellipses, in our current implementation C = 2. When an ellipse is detected the points located on its contour are not deleted from the search array, because this significantly lowers the performance of the algorithm –e.g. suppose a false ellipse is detected intersecting the real ellipse because of noise, removing its edge points would mean that we would also have to remove edge points that are present on the contour of the real ellipse, thus reducing its quality and its chances of being detected. Further filtering methods are introduced to filter out false ellipses. We only consider an ellipse to be valid is it has points distributed on both sides of the major axis, furthermore we check to see if the number of points is proportional on either side of the ellipse (when we refer to the side of the ellipse we mean the points located on its one side of its contour).

To improve ellipse detection, the accumulator is quantized, resulting in a thicker ellipse contour. Having more points on its contour, the digitization problem is overcome, because a more complete elliptic contour can be approximated with formulas (3) to (8). Due to the fact that after thresholding the original image the contours of the ellipses are thick, for one real ellipse many ellipses appear to be in the same place with slightly different parameters, so the result of the detection phase must be clustered, to obtain the real ellipses. Because we have no a priori knowledge of how many real ellipses are in the source image, standard clustering techniques such as LVQ or K-means can not be applied because they are highly dependent on the initial number of clusters. Our approach is similar to VQ. We calculate a similarity distance between two ellipses using ellipse feature vectors. The feature vectors consist of the following:

$$V(x_0, y_0, a, b, \alpha) \quad (9)$$

The measured distance is the Euclidian distance in this five dimensional feature space. The distance is calculated using the following formula:

$$D(V,W) = \sqrt{\sum_{i=1}^{5} (p_{iv} - p_{iw})^2} \quad (10)$$

where $V(p_{1v}, p_{2v}, p_{3v}, p_{4v}, p_{5v})$ and W are vectors of the type (9).

If the distance is above a certain similarity threshold – which is experimentally chosen (we used $D_T=20$) – the ellipse is considered to be different from the compared one, otherwise the two compared ellipses are said to match. Each time a different ellipse is found a new cluster is formed. Using a single pass algorithm all the detected ellipses are assigned to the clusters to which they belong. In the comparison process the centroid of the cluster is used as the representative ellipse of the group, but when outputting the result the ellipse that is most near to the centroid of the group is considered to be the detected ellipse.

*3.2 Ellipse detection algorithm results*

The performance of the algorithm is demonstrated using real world images. We present four cases of real ellipse detection either with or without foreign objects (that represent noise for the ellipse detector).

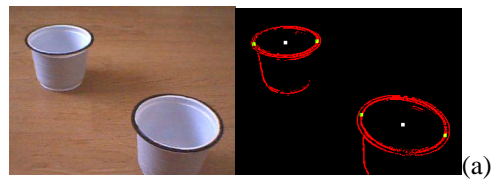
(a)

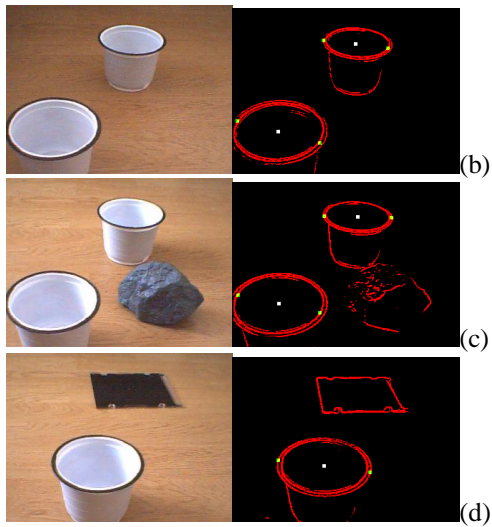

Fig. 9. Experimental Results.

Output of the ellipse detector can be seen in table 2.

TABLE 2: DETECTED ELLIPSES PARAMETERS

| Fig. | Center x (pixels) | Center y (pixels) | Major Axis (pixels) | Minor Axis (pixels) | Alfa (rad) |
|---|---|---|---|---|---|
| 9(a) | 235.5 | 164.5 | 60 | 37 | 0.253 |
| 9(a) | 71.0 | 49.0 | 44 | 22 | 0.045 |
| 9(b) | 65.5 | 174.5 | 59 | 40 | 0.276 |
| 9(b) | 173.0 | 51.5 | 45 | 22 | 0.122 |
| 9(c) | 66.0 | 173.5 | 58 | 43 | 0.222 |
| 9(c) | 176.0 | 50.0 | 47 | 22 | 0.021 |
| **9(d)** | 129.0 | 155.5 | 64 | 40 | 0.133 |

Output of the post processing phase can be seen in table 3. Ellipse quality represents the threshold of a high pass filter used to filter out false ellipses – we refer to ellipse quality meaning the number of points on its contour.

TABLE 3: DETECTED ELLIPSES STATISTICS

| Fig. | Virtual Ellipses | Real Ellipses | Ellipse Quality | Search Point Pairs | Total Edge Points |
|---|---|---|---|---|---|
| 9(a) | 48 | 2 | 200 | 6428 | 3214 |
| 9(b) | 42 | 2 | 200 | 6530 | 3265 |
| 9(c) | 16 | 2 | 230 | 8198 | 4099 |
| **9(d)** | 44 | 1 | 300 | 6238 | 3119 |

## 4. CONCLUSIONS

The robotic system proved to be a robust system which can be successfully used in academic research – e.g. multi agent research, human machine interaction. The communication can easily be upgraded to wireless communication. More powerful and resourceful embedded platforms can be used, to provide local intelligent behavior, making the system completely or partially autonomous. The ellipse detection method is applicable in real time applications, native implementations (both in hardware and software) can be very fast. It is also robust regarding false ellipse detection. It classifies correctly the detected ellipses as being similar or different. As further improvements, a more advanced adaptive method for clustering is desired, and to also improve ellipse detection and filtering using context information. The search step can be pseudo-random, ellipse positions could be predicted using local and global prediction tables or neural networks